\documentclass[10pt, journal]{IEEEtran}
\usepackage{amsmath,amsfonts}
\usepackage{amsmath}
\usepackage{amssymb}
\usepackage{xfrac}
\usepackage{soul}
\usepackage{multirow}

\usepackage[ruled,linesnumbered]{algorithm2e}
\usepackage{algpseudocode}
\usepackage{textcomp}
\usepackage{stfloats}
\usepackage{verbatim}
\usepackage{graphicx}
\usepackage{cite}
\usepackage[numbers,sort&compress]{natbib}
\usepackage{array}
\usepackage{bm}
\usepackage{balance}

\usepackage{color,xcolor}
\usepackage{booktabs}
\usepackage{array}
\usepackage{float}
\usepackage{makecell}
\usepackage{tabularx}
\usepackage{pifont}
\usepackage{circledsteps}

\makeatletter

\newcommand{\figref}[1]{Fig.\ref{#1}\;}
\newcommand{\tabref}[1]{Table \ref{#1}}
\renewcommand{\algref}[1]{Algorithm \ref{#1}}

\soulregister{\cite}7 
\soulregister{\citep}7 
\soulregister{\citet}7 
\soulregister{\ref}7 
\soulregister{\pageref}7 
\soulregister{\figref}7
\soulregister{\tabref}7

\newcommand\uhl[2][\usrcolor]{\sethlcolor{#1}\hl{#2}}

\begin{document}

\bstctlcite{IEEEexample:BSTcontrol}
\title{\uhl{DexiTac: Soft Dexterous Tactile Gripping}}

\author{
        \IEEEauthorblockN{Chenghua~Lu$^{1}$, Kailuan~Tang$^{2}$, Max Yang$^{1}$, Tianqi Yue$^{1}$, Haoran Li$^{1}$, Nathan F. Lepora$^{1}$}
        \thanks{This work was supported by a Leverhulme Trust Research Leadership Award `A biomimetic forebrain for robot touch' (RL-2016-39).}
        \thanks{\IEEEauthorblockA{$^{1}$Chenghua Lu, Max Yang, Tianqi Yue, Haoran Li and Nathan F. Lepora are with the School of Engineering Mathematics and Technology, and Bristol Robotics Laboratory, University of Bristol, Bristol, U.K. Email: {\rm\footnotesize chenghua.lu@bristol.ac.uk; max.yang@bristol.ac.uk; tianqi.yue@bristol.ac.uk;
        haoran.li@bristol.ac.uk; n.lepora@bristol. ac.uk}.}}
        \thanks{\IEEEauthorblockA{$^{2}$Kailuan Tang is with the Department of Mechanical and Electrical Engineering, Harbin Institute of Technology, Harbin 150000, China. Email: {\rm\footnotesize tangkl@mail.sustech.edu.cn}.}}
        }

\maketitle

\begin{abstract}
 Grasping objects—whether they are flat, round, or narrow and whether they have regular or irregular shapes—introduces difficulties in determining the ideal grasping posture, even for the most state-of-the-art grippers. In this article, we presented a reconfigurable pneumatic gripper with fingers that could be set in various configurations, such as hooking, supporting, closuring, and pinching. Each finger incorporates a dexterous joint, a rotating joint, and a customized plug-and-play visuotactile sensor, the DigiTac-v1.5, to control manipulation in real time. We propose a tactile kernel density manipulation strategy for simple and versatile control, including detecting grasp stability, responding to disturbances and guiding dexterous manipulations. We develop a double closed-loop control system that separately focuses on secure grasping and task management, demonstrated with tasks that highlight the capabilities above. The gripper is relatively easy to fabricate and customize, offering a promising and extensible way to combine soft dexterity and tactile sensing for diverse applications in robotic manipulation.

\end{abstract}


\IEEEpeerreviewmaketitle
\section{Introduction}

\begin{figure}[t]
  \centering
  \includegraphics[width=3.1in]{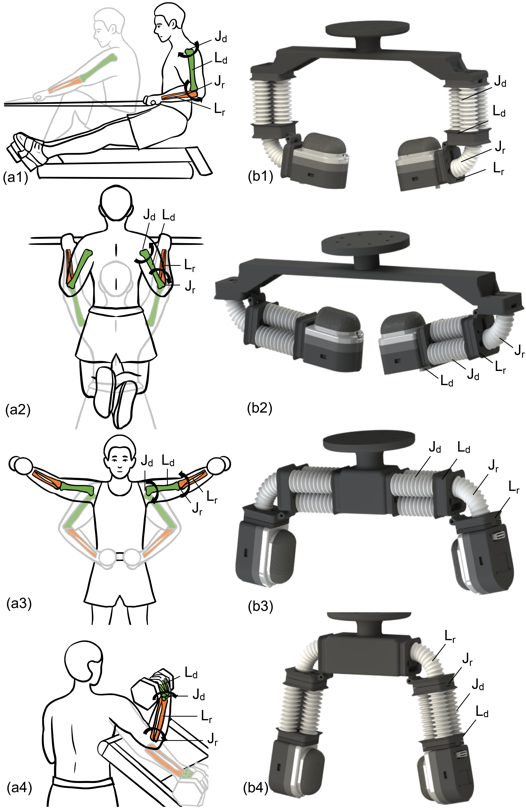}
  \caption{Design concept for the proposed reconfigurable gripper. \uhl{(a) Fitness exercises include (a1) vertical row, (a2) pull-up, (a3) dumbbell side lift and (a4) dumbbell bicep curl.} (b) The reconfigurable gripper inspired by (a1-a4) includes (b1) HookGrip, (b2) SupportGrip, (b3) ClosureGrip and (b4) PinchGrip. Jd, Jr, Ld and Lr represent the dexterous joints, rotating joints, dexterous links and rotating links, respectively.}
  \label{fig_concept}
\end{figure}

\IEEEPARstart{H}{uman} activities encompass interactions with an extensive range of objects. While this comes naturally to us, it presents significant challenges for designing robotic grippers that aim to emulate or complement human-like capabilities~\cite{sun2021research}. \uhl{There have been huge advances in the development of robotic grippers, catering for the increasingly diverse needs of various applications such as agriculture, domestic services, and manufacturing assembly lines~\cite{gripperneed}.  Traditional grippers are engineered for specialized tasks with fixed designs, and are not intended to be modified once manufactured, which restricts their versatility~\cite{nishimura2023lightweight,ottonello2023design}.} This limitation has paved the way for the rise of reconfigurable grippers. These new grippers can modify their structure according to the task, amplifying their adaptability and broadening their operational domain \cite{borisov2022reconfigurable,zhang2022pneumatically,cheng2022reconfigurable}. 


\uhl{While the reconfigurability of the gripper broadens its dynamic gripping capabilities, it also introduces new complications. Due to the use of soft materials to enhance safety and adaptability, there may be difficulties in assessing the success of a grasp during physical tasks~\cite{safe}, which could inadvertently result in damage to either the gripper or the objects being held \cite{su2020high}. Therefore, it is important to develop reconfigurable grippers equipped with integrated tactile sensing ability.\\
\indent Many tactile sensors have been developed for integration with robotic grippers, enabling novel perception capabilities. Lu et al. developed GTac-Gripper with a multimodal piezo-resistive and magnetic sensor, achieving in-hand object translation and rotation \cite{GTacGripper}. Park et al. designed a gripper with magnetic MagTac sensors capable of measuring the six-axis contact force/torque when grasping \cite{MagTac}. Grover et al. integrated Takktile Barometric tactile sensors on a Robotiq gripper and performed slip detection \cite{Takktile}. Oliver et al. developed a gripper with arrayed light vectors tactile sensors for a simple manipulation task like rudimentary lift and rotate tasks \cite{Oliver1, Oliver2}. However, these sensors, utilizing low-resolution and array-based technologies, are limited in their ability to extract high-level features that are more directly related to the characteristics of the object being held.}

\uhl{Image-based tactile sensors provide good solutions for gripper integration owing to the rich information they provide from high-resolution camera images, which make them well suited at applications like object shape reconstruction and tactile control \cite{imagebased,visionbased}. The innovative design of GelSight \cite{yuan2017gelsight} has led to the development of a range of reflection-based optical tactile sensors, such as GelSlim \cite{Slim} and DIGIT \cite{lambeta2020digit}. The TacTip family \cite{ward2018tactip} instead employs biomimetic pins and markers to amplify the contact on their surface. Their inherent compliance makes them well suited for manipulation tasks, such as grasping and accurately detecting phase changes. However, the existing sensors are customized for specific grippers, their sizes and non-modularity make them ill-fitted for reconfigurable grippers. This motivates a need for a compact, seamlessly integrated sensor with modular compatibility.}


In this article, we present a reconfigurable pneumatic soft robotic gripper that integrates modular joints and soft high-resolution tactile sensing. This design not only enables adaptability in various manipulation configurations but also ensures real-time responsiveness from the tactile feedback. The main contributions of this work are:

1) \textbf{Reconfigurable Pneumatic Gripper Design}: We propose a reconfigurable pneumatic gripper with a modular design that includes the palm, dexterous joint, rotating joint and fingertips. This allows for dynamic reshaping, enabling the gripper to adopt various grasping configurations, such as hook, support, closure or pinch grasps on a diverse set of objects.

2) \textbf{Integrated Soft High-Resolution Tactile Sensor}: We customize a new low-cost and high-resolution version of DigiTac (DigiTac-v1.5), which endows the gripper with the capability to extract rich sensory contact information, making it well-suited for a multitude of manipulating tasks.

3) \textbf{Tactile Image-Based Manipulation Strategy}: We develop a simple strategy using a kernel density transformation of tactile images for closed-loop control of the adaptable gripping. This strategy includes detecting grasping stability, responding to disturbances and guiding dexterous manipulations.

4) \textbf{Double Closed-Loop Safety \& Task Control}: We implement a real-time double closed-loop control system to ensure safe manipulation and task management. While one loop consistently monitors and adjusts the gripper's air pressure, ensuring safety, the other utilizes high-resolution tactile feedback to execute tasks efficiently.

\section{Gripper Design}
\subsection{Design Concept}

\uhl{Designing a gripper with reconfigurable ability requires the following features:

1)	High dexterity and flexibility: the gripper should have a high degree-of-freefom (DoF) to ensure versatility.

2)	Functional differentiation: different joints should have different functions to enable the gripper's various actions.

3)	High control efficiency: the combination of different joints should still simplify control.

In species with joints (e.g. mammals), a single joint's maximum DoF can be up to three~\cite{van2021exploring}, which offers the possibility of high dexterity. In the limbs of mammals, different joints undertake different functions: e.g., the shoulder joint allows complex spatial motion, whereas the elbow joint provides stability and support to give efficient rotational motion. One adaptive feature throughout long-term evolution in mammals is that the parts adjacent to high-DoF joints are typically low-DoF joints~\cite{burgess2021review}. Coordination between these joints leads to complex task performance in an efficient, economical way. 

For illustration, we examine the human upper limb movement when doing various exercises (\figref{fig_concept}). The human upper limb includes three joints: shoulder, elbow and wrist. The vertical rows, with wrists remaining stationary, rely predominantly on three-DoF in the shoulders and one-DoF in the elbow (\figref{fig_concept}(a1)). Conversely, when doing pull-ups, our body, like a payload, is supported by shoulder and elbow joints (\figref{fig_concept}(a2)). Many exercises require spreading out the shoulder joint, like a dumbbell side lift, where the wrist movement is limited and the shoulder and elbow joints maintain a large spatial motion (\figref{fig_concept}(a3)). When doing a dumbbell bicep curl, the shoulder is stationary, with only the elbow and wrists lifting and dropping the payload (\figref{fig_concept}(a4)).\\
\indent These observations motivated us to design grippers that emulate the movement patterns seen in these exercises, as highlighted in \figref{fig_concept}(b). }To boost adaptability, these were equipped with a set of tactile sensors to give a series of versatile grippers.\\

\begin{figure*}[!t]
  \centering
  \includegraphics[width=7in]{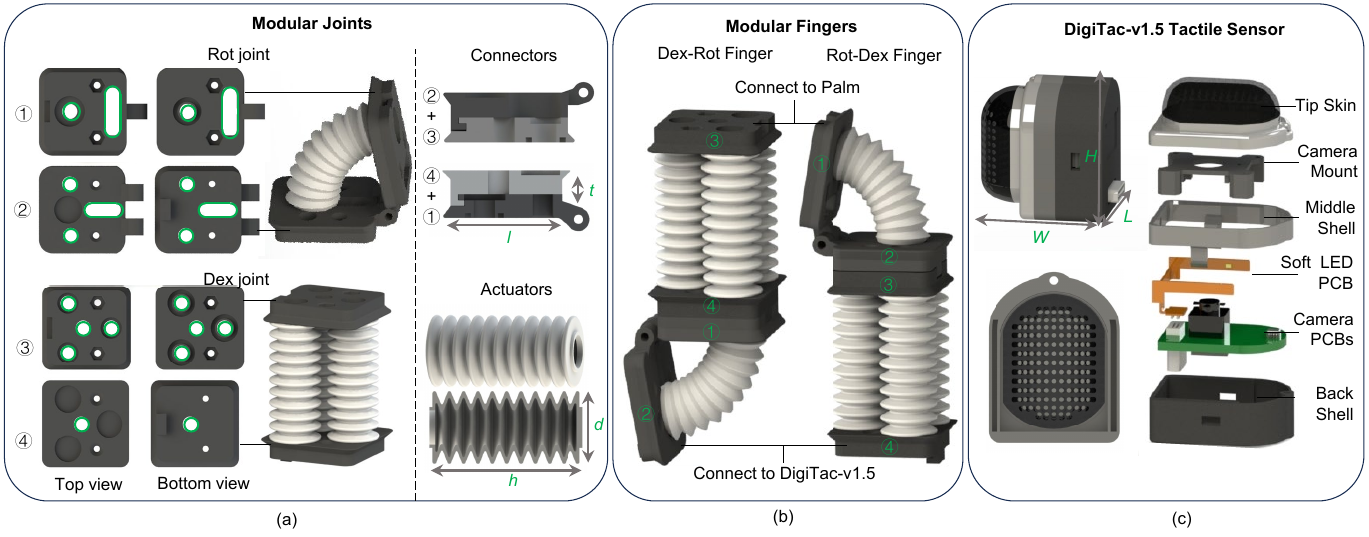}
  \vspace{-1em}
  \caption{Joints, fingers and sensor (bolts, nuts, tube excluded). (a) Rot joint, Dex joint, soft actuators and rigid connectors. Holes 
bordered in green are tube accesses and the rest are assembly holes. (b) Different assemblies of connectors and configurable fingers after assembly. (c) Overall view of DigiTac-v1.5 and its component design.}
  \label{fig_design}
\end{figure*}
\balance
\vspace{-20pt}
\subsection{Design Implementation}

Our design aims to allow many configurations by simply rearranging the assembly of palms and joints, and customizing the gripper to specific object-grasping needs and tasks. \hl{Each finger features a rotating joint (Rot joint), for fundamental flexion and extension, and a dexterous joint (Dex joint) to provide a high DoF (\figref{fig_design}(a)), so the finger has better control and lower energy consumption.}

These joints are crafted from modular bellows made of 3D-printed Polyurethane rubber (Shore hardness: $A=75$) and are complemented with modular rigid connectors crafted from 3D-printed Vero-black material with a Stratasys J826 printer. This modular approach simplifies both assembly and disassembly.



\begin{figure}[!t]
  \includegraphics[width=3.4in]{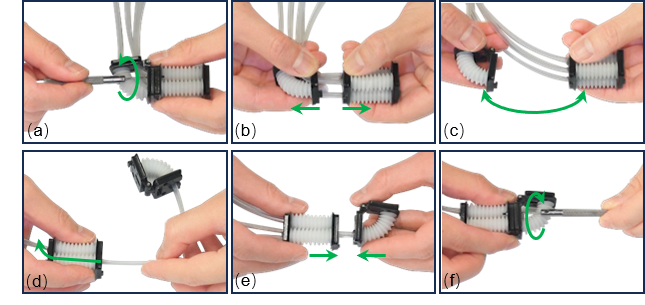}
  \caption{Reconfiguration process from Rot-Dex finger to Dex-Rot finger. (a) Unscrew the bolts. (b) Separate two joints. (c) Exchange places. (d) Enter the tube. (e) join the two joints. (f) screw the bolts.}
  \label{fig_reconfig}
\end{figure}

Two finger designs emerge depending on the assembly sequence from the base (Fig. 2(b), top) to the fingertip (Fig. 2(b), bottom): the ``Dex-Rot Finger" where the dexterous joint is first, and the ``Rot-Dex Finger" where the rotating joint is first. \uhl{The reconstruction process from Dex-Rot finger to Rot-Dex finger covers 6 steps, as shown in Fig.3.} Simply adjusting the assembly direction and pairing it with different palms allows for a range of gripper configurations, enhancing its versatility without the necessity for complete redesigns. \uhl{The reconstruction time of one finger is around $5$ minutes, the video of the process can be found in the supplementary materials. }For enhanced sensing functionality, a high-resolution tactile sensor is integrated at the gripper's tip.

\uhl{Configurations 1 and 2 employ a horizontal assembly:}

\textbf{Configuration 1 (HookGrip, Fig. 1(b1)}): With the 3-DoF Dex joint positioned at the top and the 1-DoF Rot joint below, this setup allows the finger's joint section to curve into a hook shape, suitable for tasks like hanging a bag handle. This gripper configuration is ideal for high-load grasping activities.

\textbf{Configuration 2 (SupportGrip, Fig. 1(b2)}): The 3-DoF Dex joint is oriented internally and the 1-DoF Rot joint is external. By using its body to cradle objects, this configuration offers stability during holding tasks.

\uhl{Configurations 3 and 4 employ a vertical assembly:}

\textbf{Configuration 3 (ClosureGrip, Fig. 1(b3)}): With the 3-DoF Dex joint externally positioned and the 1-DoF Rot joint internally located, this setup facilitates the encircling of sizable objects, mirroring a bat's embrace.

\textbf{Configuration 4 (PinchGrip, Fig. 1(b4)}): Placing the 1-DoF Rot joint at the top and the 3-DoF Dex joint at the bottom, the pinch grasp is the most common grasping method. This gripper configuration is highly flexible and suitable for handling cuboid objects. 

Notably, the choice of palm size (especially the length) of the gripper can provide space to ensure successful grasping. The assembly of the joints utilizes both plug-in and bolt-fastening techniques, with air pipes routed through channels (marked with green circles in Fig. 2(a)).

\subsection{DigiTac-v1.5 Design}

For improved grasping versatility and dexterity, we set various expectations on the tactile sensing capabilities:\\
\noindent   a) be of a suitable morphology and size to apply to multiple types of grasping tasks and dexterous manipulations;\\
\noindent   b) be compact and lightweight to be suitable for integrating the fingertips of soft pneumatically actuated grippers;\\
\noindent   c) be relatively low-cost, easily fabricated and versatile to apply for many applications.

In response to these requirements, we developed the DigiTac-v1.5 that implemented the following changes as an improved version of our previous DigiTac \cite{lepora2022digitac}.

\textbf{Material \& Marker Design}: Given the requirements for superior elasticity and wear resistance, we have chosen the Agilus-30 material, which is more robust and offers superior elongation at break, tear resistance and tensile strength than the prior Tango Black+ material used in DigiTac. In addition, we have updated the design of the skin surface to have semi-cylindrical and 1/4 spherical surfaces to heighten sensitivity over the flat-skinned sensors, with the cross-section area of the skin occupying about 75\% of the total volume, which is significant improvements over DigiTac (45\%) enhancing interaction with manipulated objects. 

\begin{figure*}[!t]
  \centering
  \includegraphics[width=7.14in]{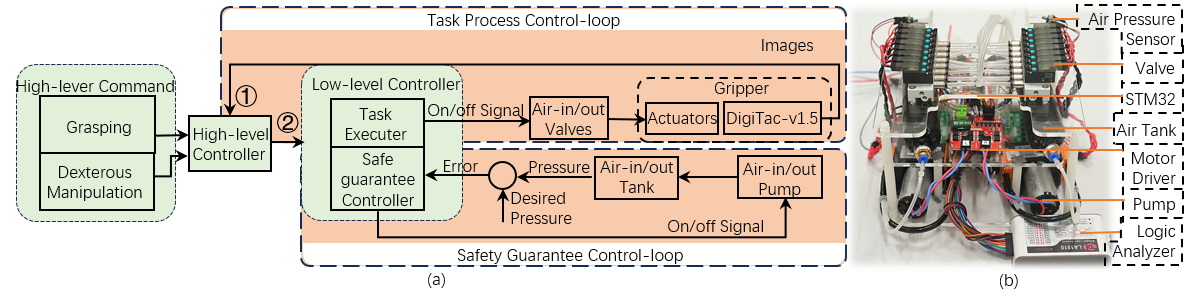}
  \vspace{-1em}
  \caption{Control System. (a) Schematic diagram of the double closed-loop control for gripper manipulation. (b) Real hardware. Two fingers' control is paralleled and achieved by multi-process, the diagram above just shows one finger's control loop as an example. More details about the multi-process control method can be found in Algorithm 1.}
  \label{fig_control_diagram}
  \vspace{-2em}
\end{figure*}

\begin{table}[t]
  \centering
  \caption{Key Geometry of the Joint}
  \label{tbl_joint_geometry}
  \renewcommand{\arraystretch}{1.25}
  \scriptsize
  \begin{tabular}[cm]{llll}
  \toprule
  $l$    Side length of the connector     &18 mm  &$h$    Length of the actuator     &26 mm\\
$t$  Thickness of the connector     &2 mm  &$d$  Diameter of the actuator     &10 mm  \\
  \bottomrule
  \end{tabular}
  \vspace{2em}
  \caption{Comparative properties of the proposed tactile sensor}
  \label{tbl_tactile_geometry}
  \renewcommand{\arraystretch}{1.25}
  \scriptsize
  \begin{tabular}[cm]{lccc}
  \toprule
   & DigiTac-v1.5 & DigiTac~\cite{lepora2022digitac} &DIGIT~\cite{lambeta2020digit}\\
  \hline
  \specialrule{0em}{1pt}{1pt}
  L$\times$W$\times$H [mm] & $22 \times 28 \times 31$  & $26 \times 39 \times 36$ & $26 \times 33 \times 36$\\
  \specialrule{0em}{1pt}{1pt}
  Weight [g] & 9.8    & 20  & 20 \\
  \specialrule{0em}{1pt}{1pt}
  \makecell[l]{2D sensing area [fraction]} & 75\% & 45\% & 45\%\\
  \specialrule{0em}{1pt}{1pt}
  \bottomrule
  \end{tabular}
\end{table}

\textbf{Camera \& Circuitry Integration}: To be compact, the most important detail is the dimension of the camera-driven board. The original DigiTac used the same VGA camera and PCB board as the DIGIT~\cite{lambeta2020digit}. In DigiTac-v1.5, we innovated a camera driver board (with the OV5693 camera, 120° view) shaped to the fingertip's contours, built on the efficient SPCA2650A chip. In addition, the camera connects to the driver using a 20-pin FPC ribbon cable, improving modularity and saving circuit board space. Using the Vero-series resin and multi-material 3D printing (Stratasys J826), the entire sensor weighs a mere 9.8\,g, halving the DigiTac's weight (Table II). 

\textbf{Assembly \& Adaptability}: To enable easy fabrication, we have transitioned from the LED board for lighting in DigiTac, to a 10-pin LED board in DigiTac-v1.5. This soft FPC ribbon cable enhances adaptability to various installation points. In addition, the Type-C port manages both power and signal connections, introducing a more usual plug-and-play feature uncommon in camera-based tactile sensing. The assembly is straightforward, culminating in an easily replicable tactile fingertip sensor (\figref{fig_control_diagram}(c)).

Overall, DigiTac-v1.5 builds on the merits of diverse camera-based tactile sensors, presenting a compact, easy-to-use, low-cost, and customizable method for tactile sensing. Hereafter, we will simply refer to the DigiTac-v1.5 as DigiTac.

\begin{figure}[!t]

  \centering
  \includegraphics[width=3.4in]{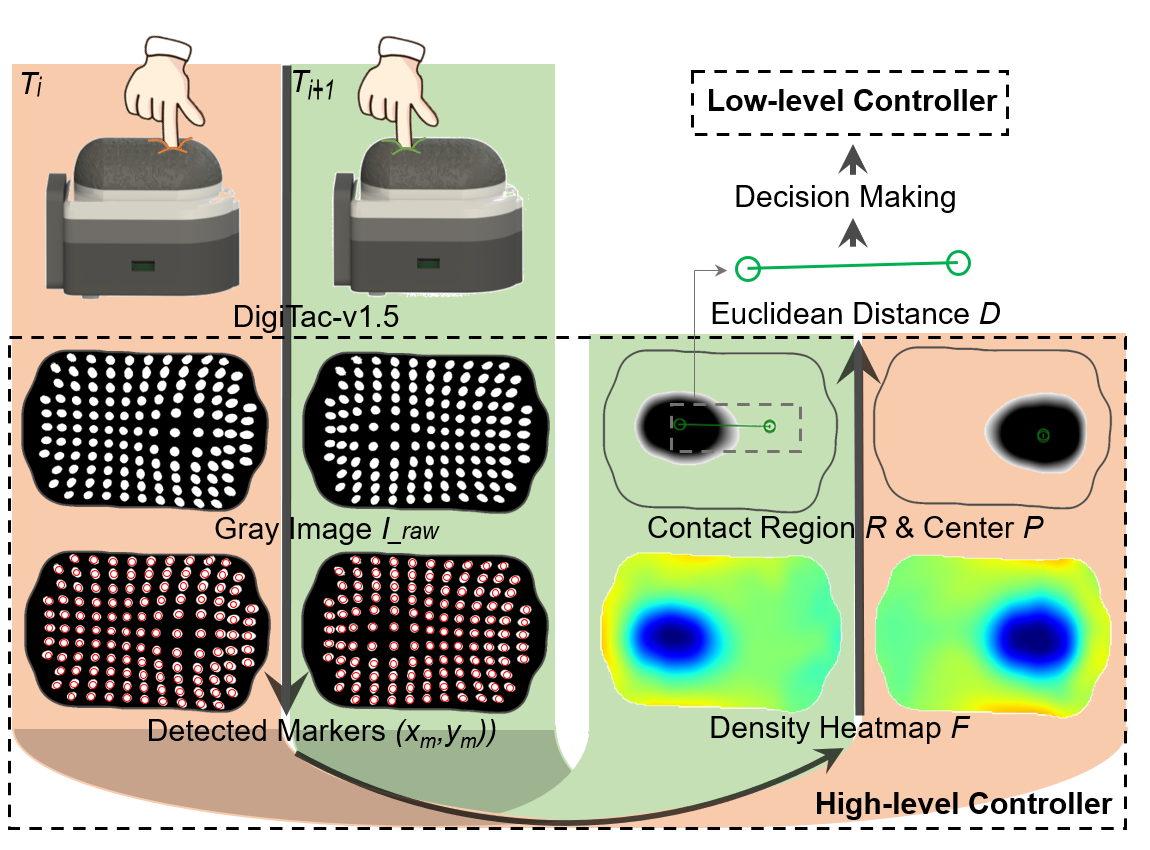}
  \caption{Image processing for DigiTac, which corresponds to steps 1-15 in Algorithm 1 (also is the details of process \ding{172} and \ding{173} in \figref{fig_control_diagram})}
  \label{fig_image_processing}
\end{figure}

\begin{algorithm}[b]
  \footnotesize
  \SetCommentSty{TimesNewRoman}
  \SetKwInOut{proca}{Process 1}
  \SetKwInOut{procb}{Process 2}
  \SetKwInOut{procc}{Process 3}
  \caption{Process for Grasping and Disturbance Response}
  \SetAlgoRefName{Algorithm 1}
  \label{alg_grasping}
  \proca{\textbf{Image processing for DigiTac1 in Finger 1}}
  \KwIn{$I_{1\_{\rm raw}}$\tcp*[r]{captured raw images from DigiTac 1}}
  \KwOut{Command Flag to Process 3}
  \While{{\rm True}}{$I_{1\_{\rm grey}}$  \tcp*[r]{cropped and rescaled gray images}
  $O_1(x_m,y_m)={\rm DOH}$  \tcp*[r]{centroids of all detected markers}
  $\bar d_1$ \tcp*[r]{estimated densities}
  $F_1$ \tcp*[r]{heatmap of the estimated densities}
  $R_1$ \tcp*[r]{the extracted contact region}
  $(x^\ast_1,y^\ast_1)$ \tcp*[r]{the extracted contact center}
  $D_1$ \tcp*[r]{distance between contact centers of consecutive frames}
  \uIf{$\rm D_1<=T_1\;\text{\rm in 3 seconds}$}{
  $\text{send}\;\rm Stale\_Grasp\;\text{to}\;Process\ 3$
  }
  \uElseIf{$\rm T_1 <D_1 <= T_2$}{
  $\text{send}\;\rm Disturbance\_Occured\; \text{to}\;Process\ 3$
  }
  \Else{$\text{send}\; \rm Regrasp\; \text{to}\;Process\ 3$}
  }
  \vspace{0.5pt}
  \hrule height .5pt
  \vspace{1.5pt}
  \procb{\textbf{Image processing for DigiTac2 in Finger 2}}
  \textbf{Similar to Process 1:} replace 1 to 2
  \vspace{0.5pt}
  \hrule height .5pt
  \vspace{1.5pt}
  \procc{\textbf{Serial communication process with low-level controller}}
  \KwIn{Flags from Process 1 and 2}
  \KwOut{Command to low-level controller}
  \uIf{{\rm Stable\_Grasp}}{
  $\text{send}\;\rm Close\_Valves\;\text{command}\;\text{to}\;MCU$
  }
  \uElseIf{$\rm Disturbance\_Occured$}{
  $\text{send}\;\rm Reopen\_Valves\;\text{command}\;\text{to}\;MCU$
  }
  \ElseIf{$\rm Regrasp$}{
  $\text{send}\;\rm Regrasp\;\text{command}\;\text{to}\;MCU$
  }
 \end{algorithm}

\section{Gripper Control}
\subsection{Hardware Control System}

A double closed-loop control system was set up to ensure the versatility of the gripper's grasping ability (see \figref{fig_control_diagram}). 
The core of the control system is an STM32F103 microcontroller unit (STM Inc.), which communicates with the high-level controller to drive the hardware during the manipulation tasks.

\begin{figure}[t]
 \centering
 \includegraphics[width=3.4in]{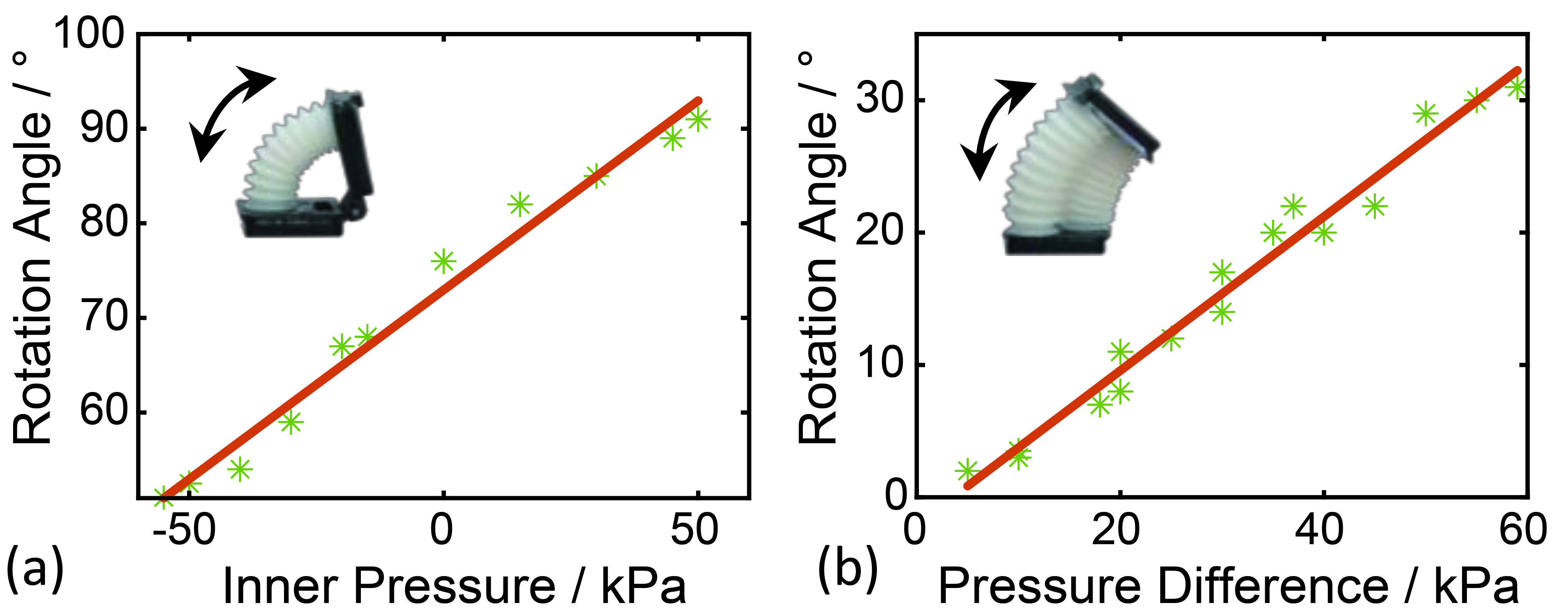}
 \caption{Relationship between joint angles and pressure. (a) Rot joint. (b) Dex joint. The pressure difference is the difference between positive pressure and negative pressure.}
 \label{fig_joint_test}
\end{figure}

The Safety Guarantee control loop consists of two diaphragm pumps (KVP8 PLUS-KK-S), which output positive and negative pressure, respectively; this uses two pressure buffer tanks (200ml, SKD Electronics) that connect to the inlet and outlet pumps, respectively, and two pressure sensors ($-100$ to $300$\,kPa, CFsensor). Instead of monitoring the pressure of each actuator, as is conventional \cite{zhu2022soft}, we simplify the pressure loop to just use two pressure sensors to measure the pressures in the tanks, ensuring the maximum and minimum pressures are always below the limit pressure of actuators (max=50\,kPa and min=$-57$\,kPa). \uhl{The pressure was obtained by using the same method in \cite{zhu2022soft} to test the characteristics of actuators, within this category, the actuator's relationship between angle and pressure input is approximately linear (see \figref{fig_joint_test})}. As a result, each joint can output a significant force without damaging the actuator, reducing the system's complexity and increasing its safety and reliability.

The Task Process control loop consists of 8 inlet and 8 outlet solenoid valves (OST Inc.) and the actuators and DigiTac sensor in the gripper. Managed by the microcontroller, the air inflates or deflates the different actuators. This allows the gripper to interact with its environment while the DigiTac captures tactile feedback, which informs the high-level controller to dictate subsequent actions.

\subsection{Control Strategy}
Based on the control method presented in the previous section, we now propose a strategy for manipulation using DigiTac tactile images for feedback. 

As the lower-level MCU activates the grasping valve, leading the fingers to open, the grasping process initiates. Concurrently, the higher-level controller monitors the tactile image feedback from the DigiTac sensor (represented in process \ding{172} of \figref{fig_control_diagram}). \uhl{As depicted in Algorithm 1 (steps 1-16)}, the raw tactile images $I_{\rm raw}$ are first transformed to grayscale images of dimension 640$\times$480 pixels. The positions $(x_m,y_m)$ of the $M$ tactile markers on the image are then detected using Determinant of Hessian (DoH) blob recognition \cite{bay2006surf} (see \figref{fig_image_processing}). \uhl{Subsequently, marker densities are estimated using a Gaussian kernel density with kernels located at marker centroids:}
\begin{equation}
    \label{eq_1}
    \!\bar d(x,y)=\frac{1}{M}\!\sum_{m=1}^{M}\!\frac{1}{\sqrt{2\pi}h^2}\exp\!\left(-\frac{||(x,y)-(x_m,y_m)||^2}{2h^2}\right)\!,\!
\end{equation}
\uhl{where $(x,y)$ represents the point at which the density is to be estimated; $||(x,y)-(x_m,y_m)||^2=(x-x_m)^2+(y-y_m)^2$ is the Euclidean distance of $(x,y)$ from the marker position; $h$ is a constant kernel width, here equal to 15 pixels which is the mean distance between adjacent markers.}

The extent of the low marker-density (blue) region in the kernel density map depends on the contact depth and gives information about the contact region. The center-of-contact $(x^\ast,y^\ast)$ is extracted as the point with the lowest density using
\begin{eqnarray}
\label{eq_2}
(x^\ast,y^\ast)={\rm arg\,min}_{(x,y)\in R}\,\bar d(x,y),
\end{eqnarray}
\uhl{where $R$ is the contact region identified as the largest contiguous region with kernel densities below a threshold value (here taking $T=0.3 /{\rm mm}^2$). It is the key contact region including the contact centre.} 


Changes in the contact center represent the interaction with the environment. Thus, we can get the trajectory of the contact area by calculating the moving Euclidean displacement that connects the contact center between frames $t$ and $t+1$
  \begin{equation}
    \label{eq_4}
    D(t)={||\left(x^\ast(t+1),y^\ast(t+1)\right)-(x^\ast(t),y^\ast(t))||},
  \end{equation}
which can be used for real-time dexterous grasp control.

To gauge the success of the grasp, we evaluate whether the contact center remains stable from whether it moves less than $T_1=0.5$\,mm (threshold 1) within a 3\,sec span. Once achieved, all pneumatic valves are sealed to maintain consistent pressure in the gripper. Subsequently, the object can be lifted, repositioned and replaced back to demonstrate a successful grasp.

\begin{figure}[!t]
  \centering
  \includegraphics[width=3.4in]{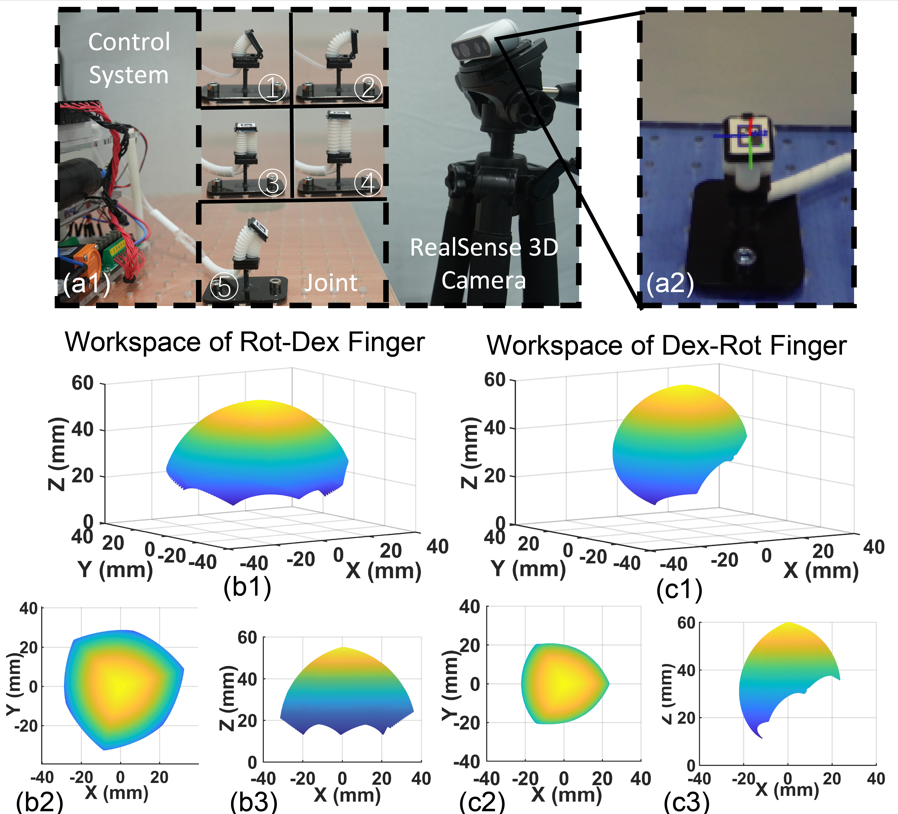}
  \caption{Workspace analysis of the modular fingers. (a) Experiment setup with (a1) hardware and (a2) view of the RealSense 3D camera. The control system is the same as Fig. 3(b). There is an ArUco marker attached to the end of the joint. (b) Workspace of Rot-Dex finger with (b1) 3D view, (b2) X-Y view, (b3) X-Z view. (c) Workspace of Dex-Rot finger with (c1) 3D view, (c2) X-Y view, (c3) X-Z view.}
  \label{fig_workspace}
\end{figure}


\begin{figure*}[!t]
  \centering
  \includegraphics[width=7in]{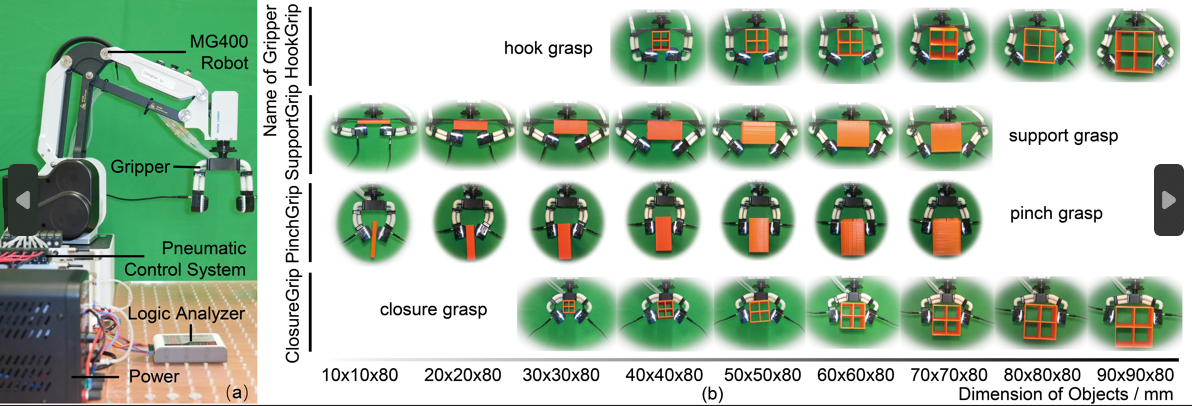}
  \caption{Setup for gripper manipulation validation experiments and grasping ability test results. (a) A desktop robot with a soft tactile gripper mounted as an end effector, alongside the pneumatics, analyzer and controller. (b) Grasping ability for different grippers with different dimensions of cuboids, showing the influence of different reconfigurable methods.}
  \label{fig_grasping_test}
  \vspace{-1em}
\end{figure*}

\begin{figure}[t]
  \centering
  \includegraphics[width=3in]{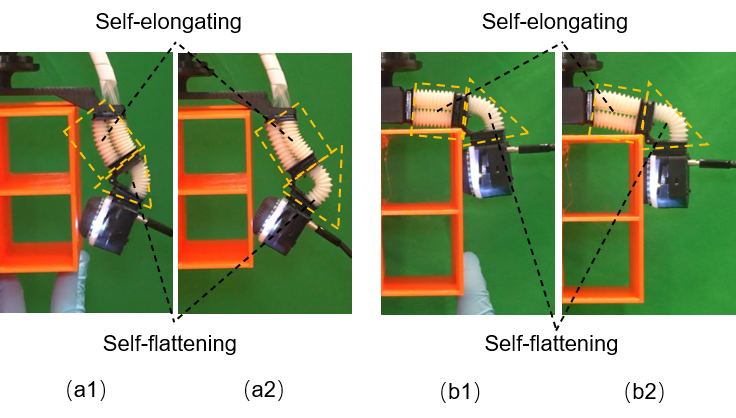}
  \vspace{-1em}
  \caption{Adaptation to the object when grasping, including the moment when HookGrip (a1) contacts and (a2) stably holds the object with dimensions of 90x90x80 mm, the moment when ClosureGrip (b1) contacts and (b2) stably holds the object with dimensions of 90x90x80 mm. }
  \label{fig_adaptation}
\end{figure}

To detect disturbances during a grasp, our approach is as follows: if the contact center shifts by more than $T_1$\,but less than $T_2=5$\,mm (threshold 2), it indicates a disturbance. Typically, such disruptions occur after the valves seal upon perceiving a successful grasp. To counteract this on detecting a disturbance, the grasping force is increased by reactivating all pertinent valves until stability is restored. However, if no contact is made with the object within 10 seconds of finger closure or if the contact center's displacement exceeds the 5\,mm threshold, it signifies a failed grasp or an object slip. In that case, the fingers release.

These control strategies are embedded within the higher-level controller as \textit{process 3}. The resulting decisions are then relayed to the lower-level systems (steps 17-23, \algref{alg_grasping}).

\section{Experimental Validation}
\subsection{Workspace Estimation}

The workspace of a gripper determines its operational range, flexibility and usability. A broader workspace enables the gripper to operate over a larger area on an object or accommodate objects of diverse dimensions. A 3D spatial distribution within the workspace equates to dexterity in the gripper's movements. Knowledge of a gripper's workspace is vital to leverage its strengths for specific grasping contexts. 

We found that each gripper configuration yielded a distinct workspace. To characterize this,  \uhl{we utilized a RealSense 3D camera (D435i, Intel Corp.) to track ArUco markers on the end of the joint (\figref{fig_workspace}(a1)), and record the marker's 3D position when the joint arrives at the limit position }(see \ding{172} to \ding{176} in \figref{fig_workspace}(a1)). \uhl{Then, we employed a constant-curvature (CC) model \cite{webster2010design} to fit the positions between the limit positions. Since each CC model of joints outputs a Denavit-Hartenberg (D-H) matrix for the kinematic calculations, we multiply the D-H matrix of the two joints in different orders to get the position of the ends of different fingers separately, following the method in previous work~\cite{tang2023strong}. }This allowed us to calculate the workspace for Dex-Rot finger and Rot-Dex finger (\figref{fig_workspace}), from which we observe: \\
\noindent a) The assembly sequence of the Rot joint and Dex joint affects the size of the workspace dimensions. The operational space of the Rot-Dex joint is far smaller than that of the Dex-Rot joint, which impacts the upper limit of the object size that can be grasped. Verification regarding this will be conducted in Sec. IV.C.\\
\noindent b) All gripper configurations exhibit a 3D spatial distribution in their workspace, implying that each gripper is capable of dexterous operations. The specific operations and their extent will be further analyzed in Sec.~IV.E.

\subsection{Experiment Setup}
To evaluate each gripper's performance, we established a custom test platform (Fig.8 (a)) comprising the gripper as the end effector affixed to a 4-DOF MG400 Desktop Robotic arm (Dobot Robotics). The DigiTacs on the gripper have a USB connection to a PC that is used to capture and store the tactile image data. The tubes of the gripper are connected to the pneumatic control system (described in Sec. III.B) powered by a 12V DC supply (NANKADF Electronics). The Logic Analyzer (Logic 16 Pro, Saleae Electronics) oversees the valve states, which relates to showing and responding to grasp stability instability (examined in Secs IV.D and E).

\begin{figure*}[!t]
  \centering
  \includegraphics[width=7in]{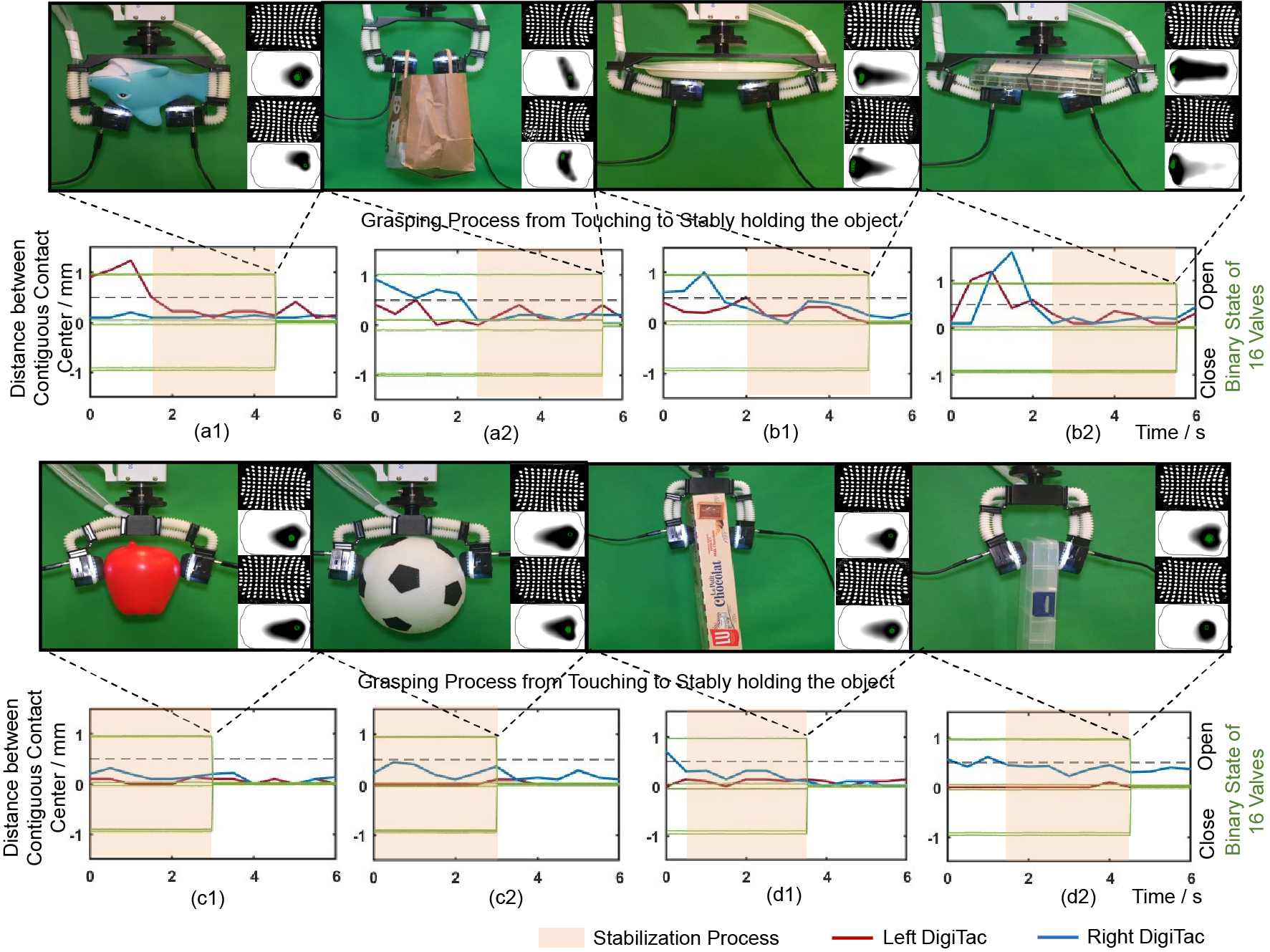}
  \caption{Representative grasping tests of each configuration and states of contact. The objects include (a1) a soft shark toy, (a2) a paper bag, (b1) a flat plate, (b2) a bolt toolbox, (c1) a round pepper, (c2) a mini soccer ball, (d1) a cookie box, (d2) a toolbox.}
  \label{fig_Representative grasping}
  \vspace{-1em}
\end{figure*}

\subsection{Grasping Adaptability Test}

\uhl{To test the grasping and its adaptability for each configuration of gripper}, we employed cuboid objects made of 3D-printed PLA material (Anycubic i3 Mega S/Pro). These objects vary in dimensions to span the workspace of each gripper, starting from a 10\,mm-sided square cross-section and increasing in 10\,mm increments up to 90\,mm, with 80\,mm length. The grasping strategy from \algref{alg_grasping} was deployed to assess each gripper's success and the resulting grasp stability. 

\uhl{The differences in the grasping are due to gripper structure, force distribution and working space. As shown in \figref{fig_grasping_test}(b), different configurations have their unique characteristics and advantages, suitable for objects of various shapes and sizes:

a) \textbf{HookGrip}: the shape of the fingers, like two hooks, can both hook onto and drag an object. The use of the Dex-Rot finger gives the Hookgrip a large working space, thereby tolerating a wider range of object sizes and shapes. During grasping, the two fingertips try to conform to the object shape, which allows the palm and fingertips to stably adapt to the surface of the object. This configuration is ideal for asymmetrical and irregularly shaped items, leveraging the large workspace.

b) \textbf{SupportGrip}: the shape of the fingers is like a tray, capable of dragging objects. Although the working space of the Rot-Dex finger is smaller, the proximity of the fingertips to the palm forms a narrow but elongated space, making it particularly suitable for gripping small or flat objects. However, when the object becomes large (here from 70-80 \,mm), the grasp will transition to a pinch grip due to the palm restricting the placement space of the object, resulting in the contact position between the fingers and the object shifting downward.

c) \textbf{PinchGrip}: the fingers oppose each other without involving the palm to hold the object. This is the most common grasping posture, and it still maintains the pinch grasp even when the grasped object becomes larger. This type of gripper is well-suited for holding objects with parallel faces.

d) \textbf{ClosureGrip}: the fingertips enclose around the object and the palm is in full or partial contact with the object. The fingers and palm come together to form a ring, attempting to envelop the object. The use of Dex-Rot makes the ClosureGrip more capable of grasping larger objects. Therefore, this type of gripper is more suitable for grasping objects with a nearly circular cross-section. For larger object sizes, the grasp again transitions into a pinch grip, like the PinchGrip above. }

When grasping larger objects, the advantages of the soft construction of the gripper become particularly evident, with \figref{fig_adaptation} showing two such examples of adaptability. To achieve a successful and stable grasp in these examples, the Dex joint elongates to accommodate the weight and size of the object; meanwhile, under constant air pressure, the Rot joint reduces its angle to adapt to the larger gripper opening angle, causing it to deform outwardly.

\begin{table}[b]
  \caption{Grasping Repeatability Test}
  \label{tab:my_label}

\scriptsize
{\begin{tabularx}{3.5in}{|X|p{0.6cm}|p{0.1cm}|p{0.3cm}|X|p{0.6cm}|p{0.1cm}|p{0.3cm}|}
  \hline
  Gripper                                                                  & Object                                                  & \multicolumn{1}{l|}{Success} & \multicolumn{1}{l|}{\begin{tabular}[c]{@{}l@{}}ATtS\\  / s\end{tabular}} & Gripper                                                                  & Object                                                  & \multicolumn{1}{l|}{Success} & \multicolumn{1}{l|}{\begin{tabular}[c]{@{}l@{}}ATtS\\  / s\end{tabular}} \\ \hline
  \multirow{2}{*}{\begin{tabular}[c]{@{}l@{}}Hook \\ Grip\end{tabular}}    & \begin{tabular}[c]{@{}l@{}}Shark \\ toy\end{tabular}    & 5/5                          & 4.23                                                                     & \multirow{2}{*}{\begin{tabular}[c]{@{}l@{}}Closure \\ Grip\end{tabular}} & \begin{tabular}[c]{@{}l@{}}Round \\ pepper\end{tabular} & 5/5                          & 3.38                                                                     \\ \cline{2-4} \cline{6-8} 
                                                                           & \begin{tabular}[c]{@{}l@{}}Paper \\ bag\end{tabular}    & 4/5                          & 4.89                                                                     &                                                                          & \begin{tabular}[c]{@{}l@{}}soccer \\ ball\end{tabular}  & 5/5                          & 3.40                                                                     \\ \hline
  \multirow{2}{*}{\begin{tabular}[c]{@{}l@{}}Support \\ Grip\end{tabular}} & \begin{tabular}[c]{@{}l@{}}Flat \\ Plate\end{tabular}   & 4/5                          & 5.03                                                                     & \multirow{2}{*}{\begin{tabular}[c]{@{}l@{}}Pinch \\ Grip\end{tabular}}   & \begin{tabular}[c]{@{}l@{}}Cookie \\ box\end{tabular}   & 5/5                          & 4.50                                                                     \\ \cline{2-4} \cline{6-8} 
                                                                           & \begin{tabular}[c]{@{}l@{}}Bolt \\ toolbox\end{tabular} & 3/5                          & 4.99                                                                     &                                                                          & \begin{tabular}[c]{@{}l@{}}Nuts \\ toolbox\end{tabular} & 4/5                          & 4.17                                                                     \\ \hline
  \end{tabularx}}
  ATtS: Average time to be stable.
  \end{table}

\subsection{Specific Grasping Capabilities}
Based on the analysis of each gripper's grasping capabilities from Sec. IV.C, we utilized each gripper to grasp household objects for which they are best suited. The processes from contacting the object to stably holding the object as well as data-capture by the tactile fingertips are shown in \figref{fig_Representative grasping}.

We use the methodology for assessing the stability of the grasp detailed in Sec.~III.B (steps 9-10 of \algref{alg_grasping}). In essence, if the shift in the contact centers across the two tactile fingertips remains below $T1$ for 3 sec, then we deem the grip to be steady.

We show tactile images of grasping to depict the moments when the grasp becomes stable in \figref{fig_Representative grasping}, and the red and blue lines on the underlying plots display the changes in the corresponding two contact centres. Overlaid on these plot are step functions that shows the binary open/close status of the 8 inflating valves and 8 deflating valves: 1 denotes the inflation valve is open, -1 indicates the suction valve is open and 0 indicates the valve is closed. The value closes when the grip is assessed as stable, with the air then sealed within the gripper.

\uhl{All grasping tests are conducted 5 times to verify the repeatability. The results are shown in the Table III.}

\begin{figure*}[!t]

  \centering
  \includegraphics[width=6.5in]{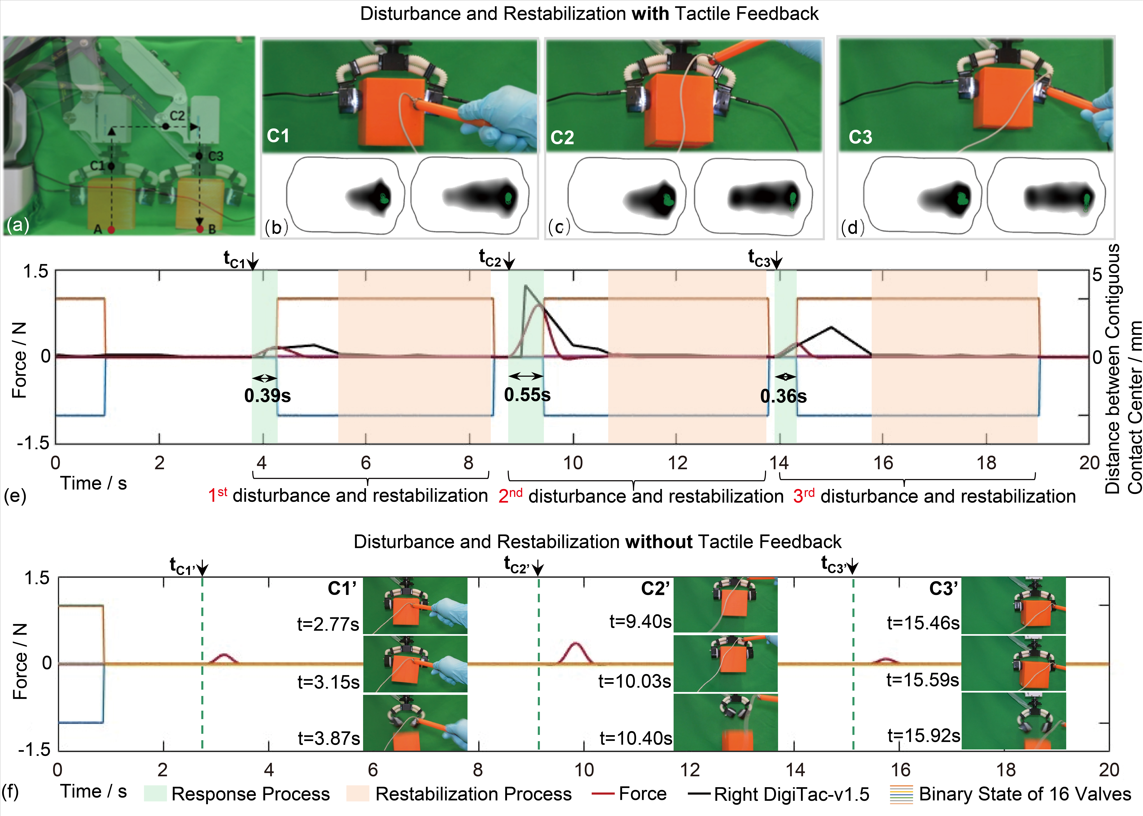}
  \vspace{-0.5em}
  \caption{Disturbance and restabilization experiment with and without tactile feedback. (a) Experiment setup. (b) Initial disturbance when poking the object. (c) Second disturbance when poking the soft actuators. (d) Third disturbance when poking the rigid back of the fingertip. Tactile image kernel densities are shown at the moments of disturbance. (e) Plots showing the disturbing force from the probe (red curve), the motion of the centres of contact (black curves) and actuators/valves response. (f) Corresponding disturbance test without tactile feedback, showing a failure to maintain grip.}
  \label{fig_disturbance_response}
  \vspace{-15pt}
\end{figure*}

\subsection{Disturbance Response}

Next, we did comparative experiments evaluating the performance with and without using tactile sensing to correct for disturbances to the grip. Here, we select one gripper configuration (ClosureGrip) as an example to assess the disturbance response ability, expecting that similar results will hold for the other gripper configurations. The experimental workflow involves grasping the object (at position A), lifting it by 120\,mm and then translating the object by 100\,mm above the endpoint (position B) where the object is lowered and the grip released. Throughout this trajectory, we manually introduced disturbances using a rod equipped with a 1-DoF force sensor (AR-DN232, Arizon Tech.) on its tip. Disturbances included:\\
\noindent (a) poking the object (\figref{fig_disturbance_response}(b));\\
\noindent (b) poking the gripper on a soft actuator (\figref{fig_disturbance_response}(c));\\
\noindent (c) poking the rigid back of a fingertip (\figref{fig_disturbance_response}(d)).

The strategy to monitor and respond to the disturbances was described in Sec.~III (Algorithm~1 steps 11-12, \uhl{20-21}). When an external force impacts the object or gripper, this force transitions to the contact interface between the object and the sensor, leading to a deviation in the center of contact. The magnitude of this shift serves as an indicator of the disturbance's influence on the grasp. If the force is identified as an external interference (using a threshold of 0.5\,mm), then measures are taken to fortify the grasp's stability. Consequently, the valves feeding to the gripper will open, closing the gripper until a stable grasping state is once again achieved (\figref{fig_disturbance_response}).

We characterize the gripper's response to disturbances using a response time calculated as the difference between when the actuators respond to when the force sensor touches the object or gripper (from a video of the experiment). This response time contains the response latency of the valves ($>$10\,ms), the delay of the high-level tactile system ($>$50\,ms) and the airspeed from the pump to the actuators (500\,mm route). Overall, this will cause a significant total response time (green background area in \figref{fig_disturbance_response}(e)). \uhl{Even so, the system could reach its fastest response time of 0.36\,sec, which is comparable to human reactions times \cite{otaki2019effect,heekeren2008neural}}.

Conversely, without tactile feedback, the gripper fails to hold the object under similar disturbances (\figref{fig_disturbance_response}(f)). The object fell or twisted then fell for these disturbances exerted on the object or gripper, which emphasizes the benefits of \algref{alg_grasping} for reacting to external disturbances.

\begin{figure*}[!t]
  \centering
  \includegraphics[width=7in]{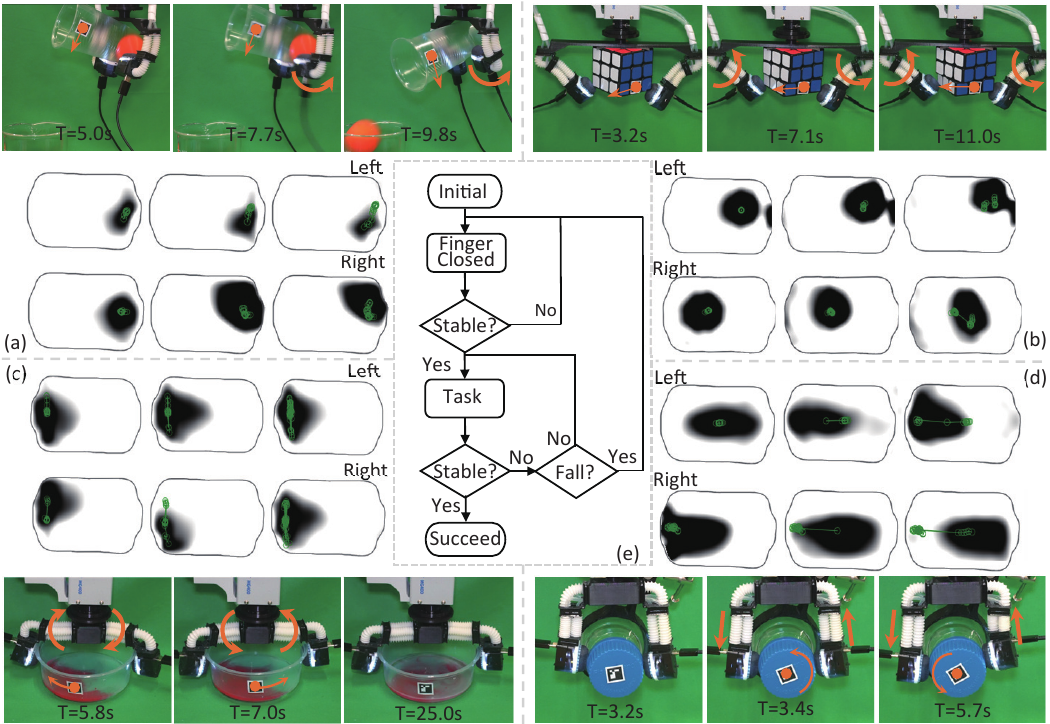}
  \caption{Dexterous manipulation with tactile sensing. (a) The HookGrip pours a ball. (b) The SupportGrip rotates a Rubik's Cube. (c) The ClosureGrip shakes the paint in the bowl until it's well mixed. (d) The PinchGrip twists a bottle's cover.}
  \label{fig_dexterous}
  \vspace{-1em}
\end{figure*}

\subsection{Dexterous Manipulation}
To indicate the dexterous capability of the grippers with tactile guiding, we give an illustrative demonstration for each configuration. \uhl{In each manipulation, tactile information identifies the status and decides the next action.} 

a) \textbf{Ball-pouring using HookGrip}. A cup containing a 30 mm diameter ball is handed to HookGrip, \uhl{then securely grasps it. Once tactile information confirms a stable hold, HookGrip proceeds to the next step: tilting backward to pour the ball into a beaker. The real-time contact data is continuously monitored during the entire process, up to the completion of the pour. \figref{fig_dexterous}(a) captures the critical moment of stable gripping and the successful release of the ball. The cup's angle and pouring trajectory are both tracked using an ArUco marker, confirming a precise pouring angle of $47.5^\circ$.}

b) \textbf{Rubik's cube-rotating using SupportGrip}. \uhl{A Rubik’s cube (size $57 \times 57 \times 57$\,mm) was put beneath the palm of the gripper. Once the gripper achieved stable contact, the gripper began to rotate the cube's bottom face, which was observed through tactile images. We conducted this rotating procedure twice. \figref{fig_dexterous}(b) illustrates key moments: the initial contact with the cube, the completion of the first rotation, and the completion of the second rotation. Notably, the contact areas on the sensor during the first rotation did not overlap with those during the second rotation, demonstrating the sensor's sensitivity.} An ArUco marker was utilized to track the rotation angle, which averaged $14.5^\circ$ per rotation.

c) \textbf{Dye-mixing using ClosureGrip}. A bowl containing 10\,ml of water is handed to the gripper. Subsequently,  1.5\,ml of red and 1.5\,ml of green dye are quickly added to the bowl. \uhl{Once stable grasping was confirmed via tactile information, the gripper began to rapidly rotate clockwise and counterclockwise while also swaying horizontally left and right. Since these actions primarily cause movement of tactile information near the $y$-axis, any significant displacement in the $x$-direction would suggest that the bowl is at risk of falling. The tactile information is continuously monitored in real-time throughout the process, with a focus on the $x$-value of the contact center's position. The movement was tracked using an ArUco marker, which recorded an average rotation angle of $\pm22.6^\circ$ and an average horizontal sway of $\pm 13$\,mm (\figref{fig_dexterous}(c)).}

d) \textbf{Cap-twisting using PinchGrip}. A bottle cap of diameter 50\,mm is placed vertically in the gripper. \uhl{After a stable hold on the cap, the gripper fingers were rotated to twist the cap until center of contact approached the edge of the tactile sensor (trajectory marked on \figref{fig_dexterous}(d)). }Then the gripper was commanded to stop twisting and return to the original grasping position, after which the cap could be twisted again. An AruCo marker was again used to monitor the trajectory, measuring a maximum twisting angle $25.2^\circ$.
   
\begin{table*}[t]
  
  \colorbox{white}{
  \begin{minipage}{\textwidth}
  \centering
  \caption{Comparison of the proposed gripper with the existing state-of-the-art grippers}
    
  \label{Tab3}
  \renewcommand{\arraystretch}{1.25}

   \begin{tabular}{@{}p{2.5cm} ccccccc p{1.7cm} p{1.7cm} p{1.7cm} p{1.7cm} p{1.7cm} p{1.7cm} p{1.7cm}  @{}}
  \toprule
Grippers   & This Work & \cite{RUTH}     & \cite{james2021tactile}    & \cite{Sensorized}  & \cite{ben2016M2}   &\cite{ben2017G2}   &\cite{GTacGripper}    \\
  \hline
  Stiffness   & Soft-rigid  & Soft-rigid   & Soft-rigid  & Soft   & Soft-rigid & Rigid   & soft-Rigid                    \\
  \hline
  Actuation System  & Pneumatic  & Tendon   &Tendon   & Pneumatic   &Tendon  & Motor  & Motor        \\
  \hline
  DOFs of Finger    & 4   & 1   & 2   & 2  & 2  & 2  & 2                                                      \\
  \hline
 DOFs of Actuation    & 2   & 3   & 4   & 3  & 2  & 2  & 8                                                      \\
  \hline
  Sensing Principle  & Image-based     & -   &Image-based   & Piezoresistive  & Image-based    & Image-based   & Piezoresistive                                  \\
  \hline
  Reconfigurations    & 4 & 3  &-   & 3    & -     &-  & 5                                           \\
  \hline

  Dexterity& 
  \begin{tabular}[c]{@{}c@{}}
    Bend, Twist,\\Extend/Contract\\Rotate,
  \end{tabular}    & Bend  & Bend     & Bend    & Bend    & Bend   &   \begin{tabular}[c]{@{}c@{}}
    Bend, Twist
  \end{tabular}    \\
  \bottomrule
  \end{tabular}
  \end{minipage}
  }
  \end{table*}

\section{Conclusion and Future Work}
In this article, a two-finger, easy-to-fabricate, biomimetic, reconfigurable pneumatic gripper with tactile sensing ability was proposed, which was inspired by the different movements of fitness exercises. With different configurations like hook-, support-, closure- and pinch-grips, the gripper had distinct workspaces for the two different `Rot-Dex'`and `Dex-Rot' finger configurations. Overall, this results in various grasping capabilities from holding flat to round shapes across a range of sizes. Grippers with Rot-Dex fingers are suited for grasping larger objects, while grippers with Dex-Rot finger are adept at handling small objects. With the addition of tactile sensing and monitoring/control via the inferred centre of contact, we demonstrated that these grippers can undertake real-time grasping stability detection and dexterous manipulation. \uhl{This includes grasp stability under external disturbances (response time $\sim$0.36\,sec) and versatile manipulations such as pouring, rotating, shaking and twisting held objects (up to $47.5^\circ$ in on motion). The summary of the performance of the proposed gripper is shown in Table IV with a comparison to other gripper designs.

In future extensions of this work, we could promote the design of the fingertip to strengthen the grasping ability, extend the control method to execute variable grasping stiffness adjustment, extract more information from the contact image and adapt a learning-based method to do recognition for environments such as the hardness, texture and classification of the objects.} Another direction would be to build multi-fingered hands to improve the gripper's dexterous manipulation capabilities of held objects. \uhl{We would also like to see our gripper applied to wide range of fields like agriculture and horticulture, waste sorting and recycling, and domestic service robots.}

\section*{Acknowledgment}
The authors would like to thank Andrew Stinchcombe, Ugnius Bajarunas and Jason Welsby of Bristol Robotics Laboratory for their engineering assistance in this research.


\bibliographystyle{IEEEtran_0}
\small{\bibliography{references/reference.bib}}

\begin{thebibliography}{10}
\providecommand{\url}[1]{#1}
\csname url@samestyle\endcsname
\providecommand{\newblock}{\relax}
\providecommand{\bibinfo}[2]{#2}
\providecommand{\BIBentrySTDinterwordspacing}{\spaceskip=0pt\relax}
\providecommand{\BIBentryALTinterwordstretchfactor}{4}
\providecommand{\BIBentryALTinterwordspacing}{\spaceskip=\fontdimen2\font plus
\BIBentryALTinterwordstretchfactor\fontdimen3\font minus \fontdimen4\font\relax}
\providecommand{\BIBforeignlanguage}[2]{{%
\expandafter\ifx\csname l@#1\endcsname\relax
\typeout{** WARNING: IEEEtran.bst: No hyphenation pattern has been}%
\typeout{** loaded for the language `#1'. Using the pattern for}%
\typeout{** the default language instead.}%
\else
\language=\csname l@#1\endcsname
\fi
#2}}
\providecommand{\BIBdecl}{\relax}
\BIBdecl

\bibitem{sun2021research}
Y.~Sun \emph{et~al.}, ``Research challenges and progress in robotic grasping and manipulation competitions,'' \emph{IEEE robotics and automation letters}, vol.~7, no.~2, pp. 874--881, 2021.

\bibitem{gripperneed}
S.~E. Navarro \emph{et~al.}, ``Proximity perception in human-centered robotics: A survey on sensing systems and applications,'' \emph{IEEE Transactions on Robotics}, vol.~38, no.~3, pp. 1599--1620, 2022.

\bibitem{nishimura2023lightweight}
T.~Nishimura \emph{et~al.}, ``Lightweight high-speed and high-force gripper for assembly,'' \emph{IEEE/ASME Transactions on Mechatronics}, 2023.

\bibitem{ottonello2023design}
E.~Ottonello \emph{et~al.}, ``Design and validation of a push-latch gripper made in additive manufacturing,'' \emph{IEEE/ASME Transactions on Mechatronics}, 2023.

\bibitem{borisov2022reconfigurable}
I.~I. Borisov \emph{et~al.}, ``Reconfigurable underactuated adaptive gripper designed by morphological computation,'' in \emph{2022 International Conference on Robotics and Automation (ICRA)}.\hskip 1em plus 0.5em minus 0.4em\relax IEEE, 2022, pp. 1130--1136.

\bibitem{zhang2022pneumatically}
Z.~Zhang \emph{et~al.}, ``Pneumatically controlled reconfigurable bistable bionic flower for robotic gripper,'' \emph{Soft Robotics}, vol.~9, no.~4, pp. 657--668, 2022.

\bibitem{cheng2022reconfigurable}
P.~Cheng \emph{et~al.}, ``Reconfigurable bionic soft pneumatic gripper for fruit handling based on shape and size adaptation,'' \emph{Journal of Physics D: Applied Physics}, vol.~56, no.~4, p. 044003, 2022.

\bibitem{safe}
Y.~Hao and Y.~Visell, ``Beyond soft hands: Efficient grasping with non-anthropomorphic soft grippers,'' \emph{Frontiers in Robotics and AI}, vol.~8, 2021.

\bibitem{su2020high}
Y.~Su \emph{et~al.}, ``A high-payload proprioceptive hybrid robotic gripper with soft origamic actuators,'' \emph{IEEE Robotics and Automation Letters}, vol.~5, no.~2, pp. 3003--3010, 2020.

\bibitem{GTacGripper}
Z.~Lu \emph{et~al.}, ``Gtac-gripper: A reconfigurable under-actuated four-fingered robotic gripper with tactile sensing,'' \emph{IEEE Robotics and Automation Letters}, vol.~7, no.~3, pp. 7232--7239, 2022.

\bibitem{MagTac}
S.~Park \emph{et~al.}, ``Magtac: Magnetic six-axis force/torque fingertip tactile sensor for robotic hand applications,'' in \emph{2023 IEEE International Conference on Robotics and Automation (ICRA)}, 2023, pp. 10\,367--10\,372.

\bibitem{Takktile}
A.~Grover \emph{et~al.}, ``Under pressure: Learning to detect slip with barometric tactile sensors,'' 2022.

\bibitem{Oliver1}
O.~Leslie \emph{et~al.}, ``A tactile sensing concept for 3-d displacement and 3-d force measurement using light angle and intensity sensing,'' \emph{IEEE Sensors Journal}, vol.~23, no.~18, pp. 21\,172--21\,188, 2023.

\bibitem{Oliver2}
O.~Leslie \emph{et~al.}, ``Design, fabrication, and characterization of a novel optical six-axis distributed force and displacement tactile sensor for dexterous robotic manipulation,'' \emph{Sensors}, vol.~23, no.~24, 2023.

\bibitem{imagebased}
K.~Shimonomura, ``Tactile image sensors employing camera: A review,'' \emph{Sensors}, vol.~19, no.~18, 2019.

\bibitem{visionbased}
U.~H. Shah \emph{et~al.}, ``On the design and development of vision-based tactile sensors,'' \emph{Journal of Intelligent \& Robotic Systems}, vol. 102, pp. 1--27, 2021.

\bibitem{yuan2017gelsight}
W.~Yuan \emph{et~al.}, ``Gelsight: High-resolution robot tactile sensors for estimating geometry and force,'' \emph{Sensors}, vol.~17, no.~12, p. 2762, 2017.

\bibitem{Slim}
E.~Donlon \emph{et~al.}, ``Gelslim: A high-resolution, compact, robust, and calibrated tactile-sensing finger,'' in \emph{2018 IEEE/RSJ International Conference on Intelligent Robots and Systems (IROS)}, 2018, pp. 1927--1934.

\bibitem{lambeta2020digit}
M.~Lambeta \emph{et~al.}, ``Digit: A novel design for a low-cost compact high-resolution tactile sensor with application to in-hand manipulation,'' \emph{IEEE Robotics and Automation Letters}, vol.~5, no.~3, pp. 3838--3845, 2020.

\bibitem{ward2018tactip}
B.~Ward-Cherrier \emph{et~al.}, ``The tactip family: Soft optical tactile sensors with 3d-printed biomimetic morphologies,'' \emph{Soft robotics}, vol.~5, no.~2, pp. 216--227, 2018.

\bibitem{van2021exploring}
J.~van Beesel \emph{et~al.}, ``Exploring the functional morphology of the gorilla shoulder through musculoskeletal modelling,'' \emph{Journal of Anatomy}, vol. 239, no.~1, pp. 207--227, 2021.

\bibitem{burgess2021review}
S.~Burgess, ``A review of linkage mechanisms in animal joints and related bioinspired designs,'' \emph{Bioinspiration \& Biomimetics}, vol.~16, no.~4, p. 041001, 2021.

\bibitem{lepora2022digitac}
N.~F. Lepora \emph{et~al.}, ``Digitac: A digit-tactip hybrid tactile sensor for comparing low-cost high-resolution robot touch,'' \emph{IEEE Robotics and Automation Letters}, vol.~7, no.~4, pp. 9382--9388, 2022.

\bibitem{zhu2022soft}
W.~Zhu \emph{et~al.}, ``A soft-rigid hybrid gripper with lateral compliance and dexterous in-hand manipulation,'' \emph{IEEE/ASME Transactions on Mechatronics}, vol.~28, no.~1, pp. 104--115, 2022.

\bibitem{bay2006surf}
H.~Bay \emph{et~al.}, ``Surf: Speeded up robust features,'' in \emph{Computer Vision--ECCV 2006: 9th European Conference on Computer Vision, Graz, Austria, May 7-13, 2006. Proceedings, Part I 9}.\hskip 1em plus 0.5em minus 0.4em\relax Springer, 2006, pp. 404--417.

\bibitem{webster2010design}
R.~J. Webster~III and B.~A. Jones, ``Design and kinematic modeling of constant curvature continuum robots: A review,'' \emph{The International Journal of Robotics Research}, vol.~29, no.~13, pp. 1661--1683, 2010.

\bibitem{tang2023strong}
K.~Tang \emph{et~al.}, ``A strong underwater soft manipulator with planarly-bundled actuators and accurate position control,'' \emph{IEEE Robotics and Automation Letters}, 2023.

\bibitem{otaki2019effect}
M.~Otaki and K.~Shibata, ``The effect of different visual stimuli on reaction times: a performance comparison of young and middle-aged people,'' \emph{Journal of physical therapy science}, vol.~31, no.~3, pp. 250--254, 2019.

\bibitem{heekeren2008neural}
H.~R. Heekeren \emph{et~al.}, ``The neural systems that mediate human perceptual decision making,'' \emph{Nature reviews neuroscience}, vol.~9, no.~6, pp. 467--479, 2008.

\bibitem{RUTH}
Q.~Lu \emph{et~al.}, ``Systematic object-invariant in-hand manipulation via reconfigurable underactuation: Introducing the ruth gripper,'' \emph{The International Journal of Robotics Research}, vol.~40, no. 12-14, pp. 1402--1418, 2021.

\bibitem{james2021tactile}
J.~W. James \emph{et~al.}, ``Tactile model o: Fabrication and testing of a 3d-printed, three-fingered tactile robot hand,'' \emph{Soft Robotics}, vol.~8, no.~5, pp. 594--610, 2021.

\bibitem{Sensorized}
J.~H. Low \emph{et~al.}, ``Sensorized reconfigurable soft robotic gripper system for automated food handling,'' \emph{IEEE/ASME Transactions on Mechatronics}, vol.~27, no.~5, pp. 3232--3243, 2022.

\bibitem{ben2016M2}
B.~Ward-Cherrier \emph{et~al.}, ``Tactile manipulation with a tacthumb integrated on the open-hand m2 gripper,'' \emph{IEEE Robotics and Automation Letters}, vol.~1, no.~1, pp. 169--175, 2016.

\bibitem{ben2017G2}
B.~Ward-Cherrier \emph{et~al.}, ``Model-free precise in-hand manipulation with a 3d-printed tactile gripper,'' \emph{IEEE Robotics and Automation Letters}, vol.~2, no.~4, pp. 2056--2063, 2017.

\end{thebibliography}


\vspace{-3em}
\begin{IEEEbiography}
  [{\includegraphics[width=0.8in,clip,  keepaspectratio]{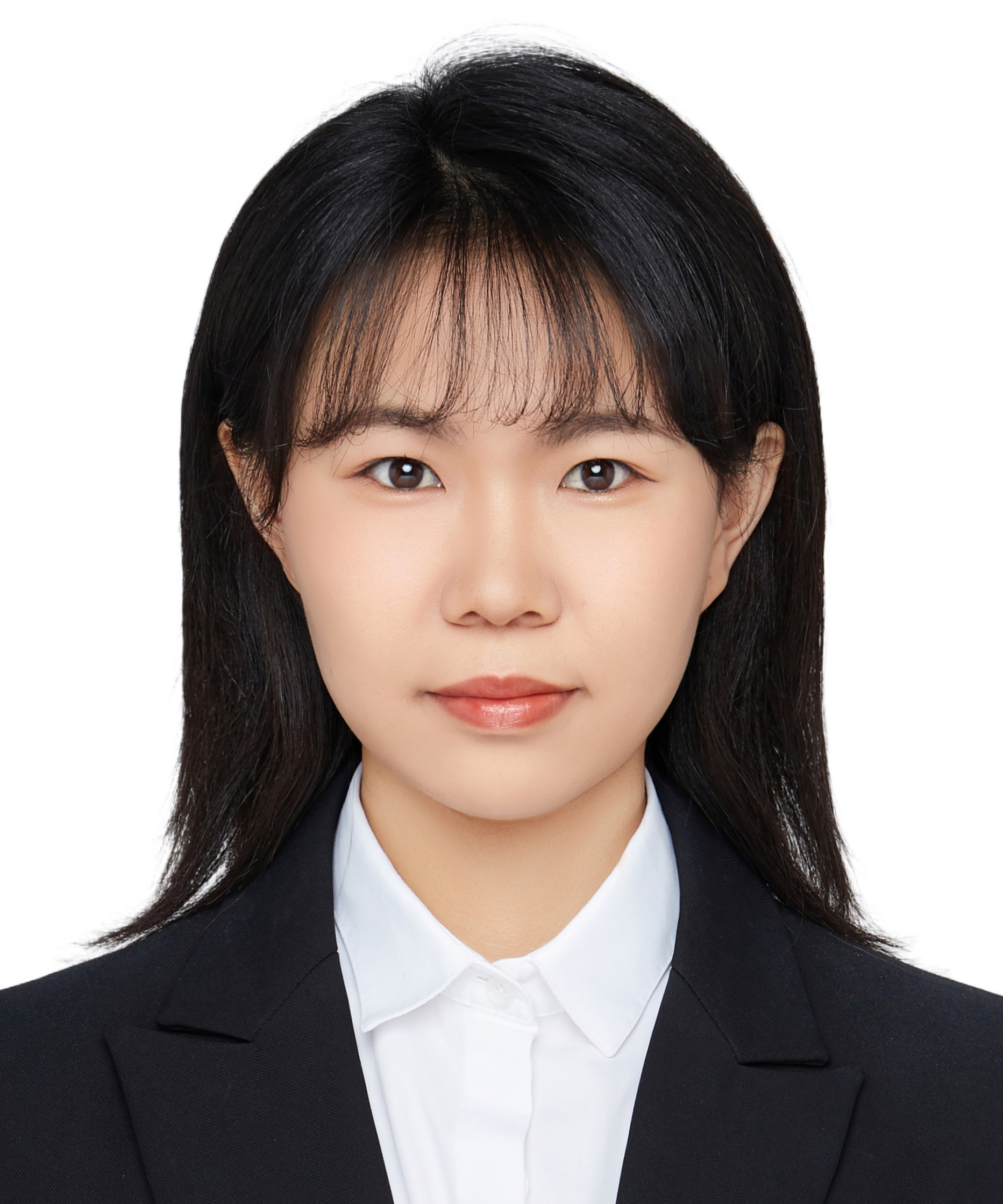}}]
  {Chenghua Lu}
  received the B.S. degree in mechanical engineering from Northeastern University, Shenyang, China, in 2017, and the M.S. degree in mechanical manufacturing and automation from the University of Chinese Academy of Sciences, Beijing, China, in 2021. She is currently working toward the Ph.D. degree majoring in Engineering Mathematics with the School of Mathematics Engineering and Technology and Bristol Robotics Laboratory, University of Bristol, Bristol, UK. Her research interests include tactile sensing and soft robotics. 
\end{IEEEbiography}

\vspace{-3em}
\begin{IEEEbiography}
  [{\includegraphics[width=0.8in,clip,  keepaspectratio]{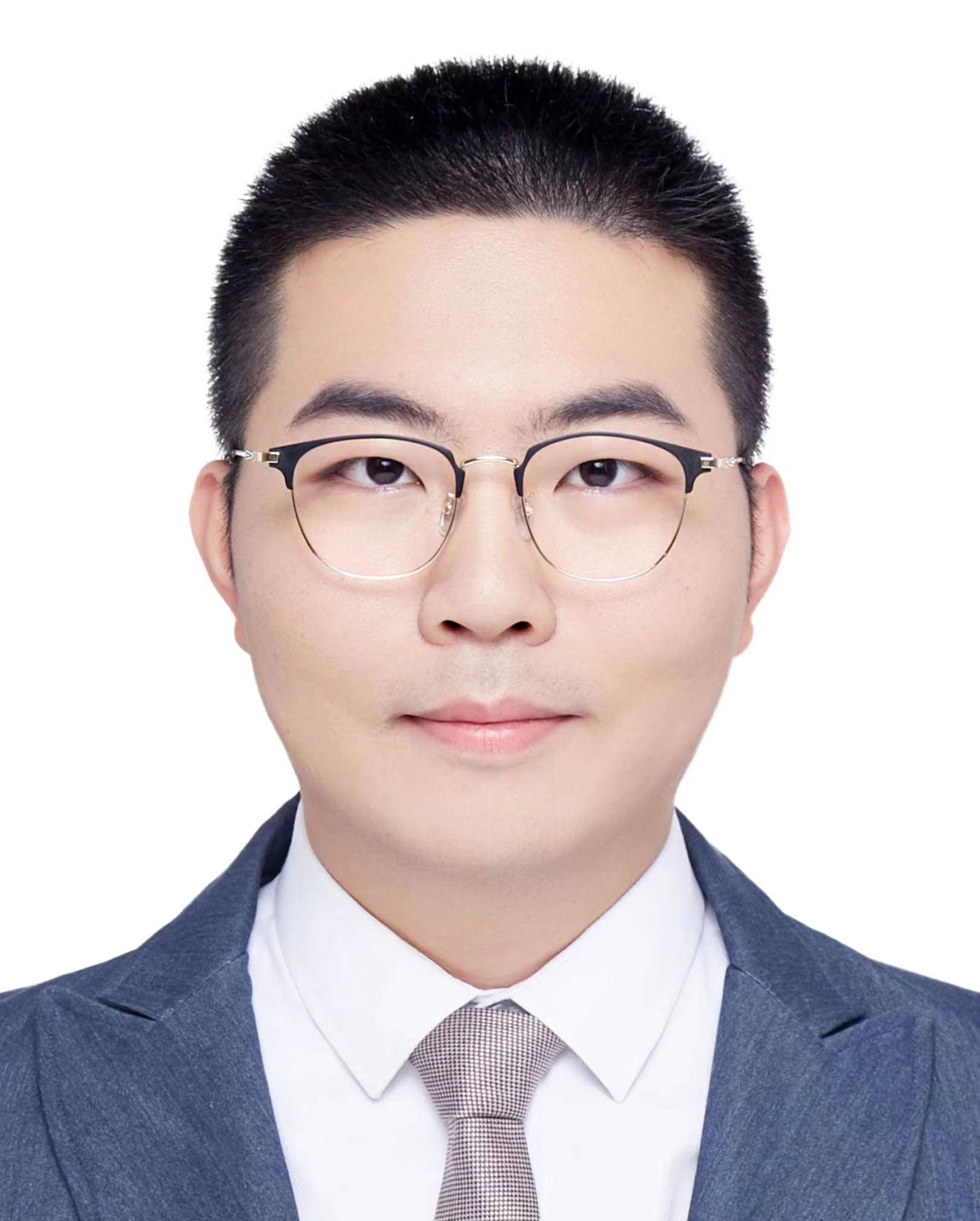}}]
  {Kailuan Tang}
  received a B.S. degree in communication engineering from the Southern University of Science and Technology (SUSTech), Shenzhen, China in 2017. He is currently working towards a Ph.D. degree majoring in mechanics with the School of Mechatronics Engineering, Harbin Institute of Technology. He is currently with the BioRobotics and Control Laboratory. His research interests include underwater robots, biomimetic control, and soft robot perception.
\end{IEEEbiography}

\vspace{-3em}
\begin{IEEEbiography}
  [{\includegraphics[width=0.8in,clip,  keepaspectratio]{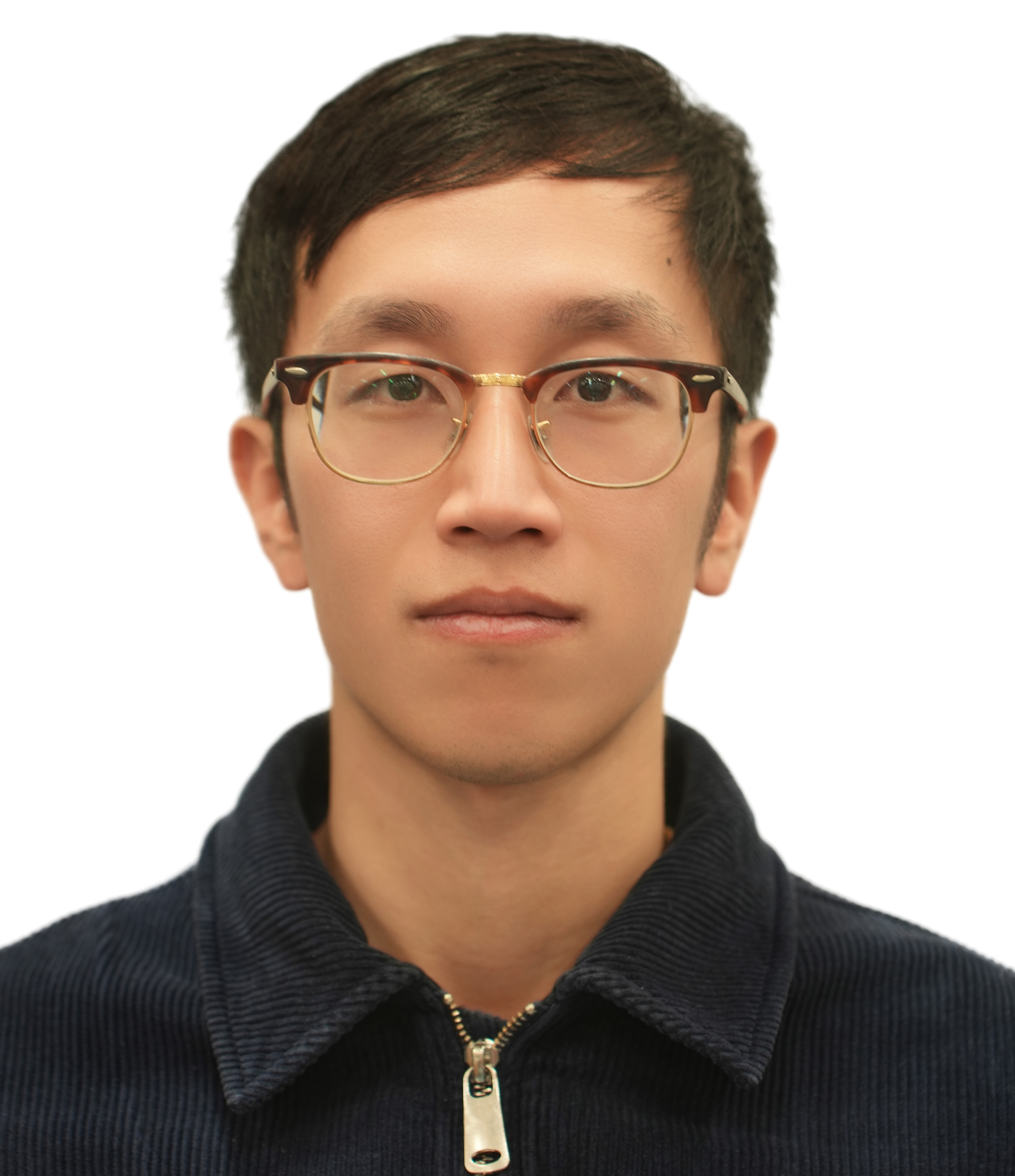}}]
  {Max Yang}
  received his MEng degree in Aeronautical Engineering from Imperial College London, UK, in 2019. He is currently a PhD student studying Robotics and AI with the Department of Engineering Mathematics and Technology at the University of Bristol and is part of the Dexterous Robotics Group at Bristol Robotics Laboratory. His research interests include machine learning, dexterous manipulation, robot control, and tactile sensing.
\end{IEEEbiography}

\vspace{-3em}
\begin{IEEEbiography}
  [{\includegraphics[width=0.8in,clip,  keepaspectratio]{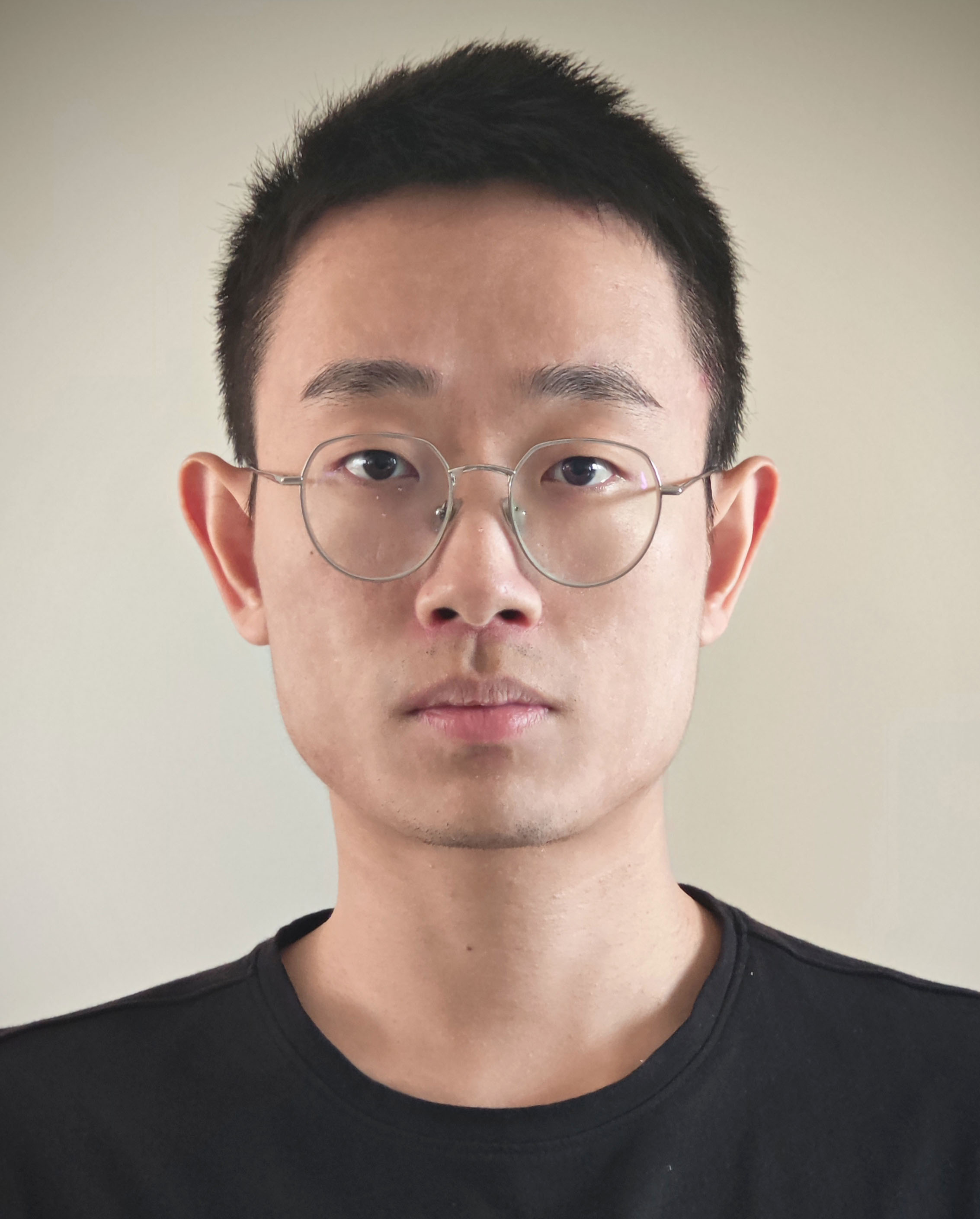}}]
  {Tianqi Yue}
  received the B.S. degree in mechanical engineering from Dalian University of Technology, Dalian, China, in 2017, and the M.S. degree in mechatronic engineering from Harbin Institute of Technology, Harbin, China, in 2019. He is currently pursuing the Ph.D. degree and working as a research associate at the University of Bristol and Bristol Robotic Laboratory. His research interests include soft robotic grippers and sensors.
\end{IEEEbiography}

\vspace{-3em}
\begin{IEEEbiography}
  [{\includegraphics[width=0.8in,clip,  keepaspectratio]{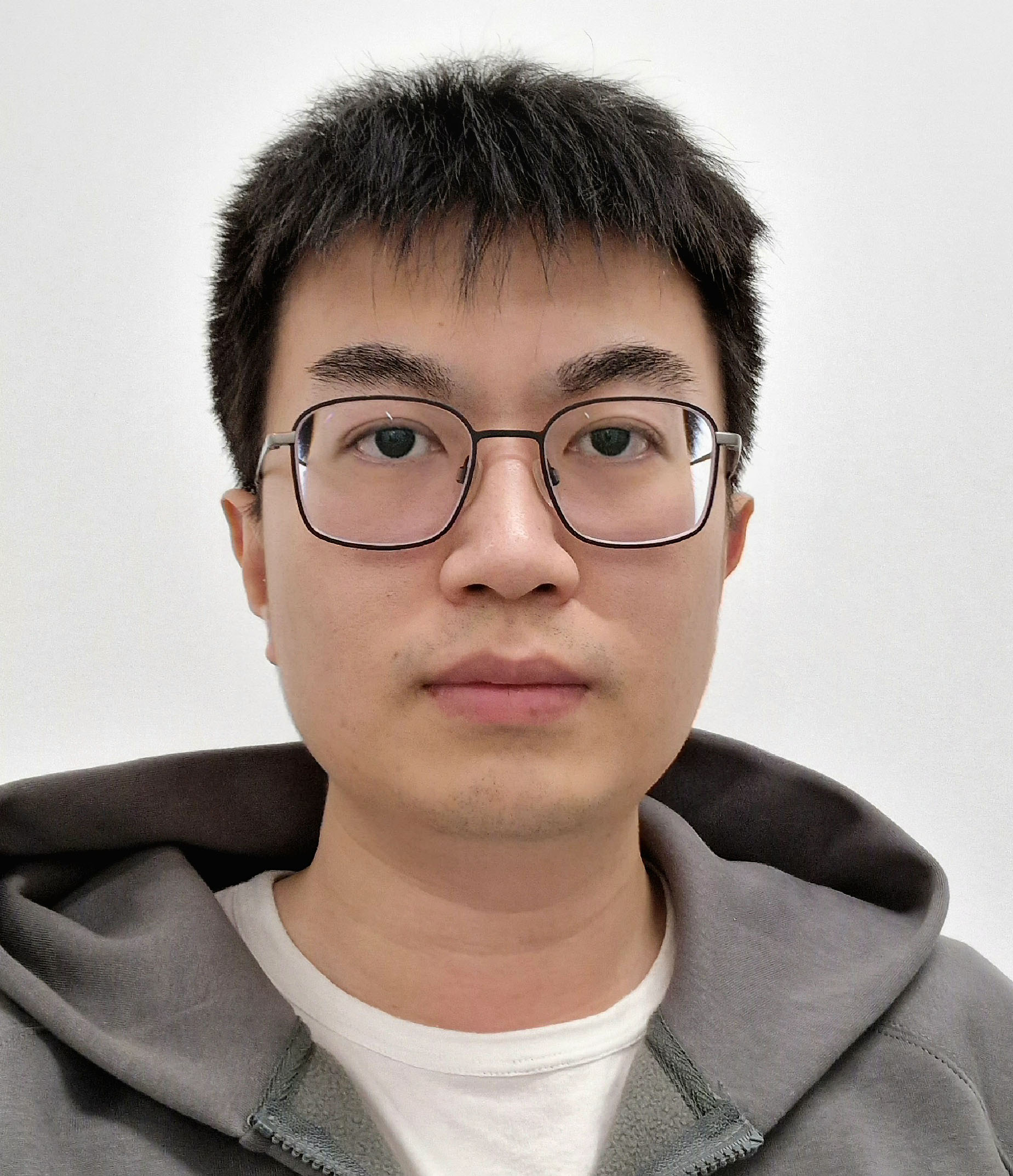}}]
  {Haoran Li}
  received B.E. in Vehicle Engineering from Wuhan University of Technology, China, in 2019, and M.Sc. degree in Robotics from the University of Bristol, in 2020. He is currently working toward the Ph.D. degree majoring in engineering mathematics with the Department of Engineering Mathematics, University of Bristol, Bristol, United Kingdom. He is currently with the Bristol Robotics Laboratory. His research interests include tactile sensors and robot hands.
\end{IEEEbiography}

\vspace{-3em}
\begin{IEEEbiography}
  [{\includegraphics[width=0.8in,clip,  keepaspectratio]{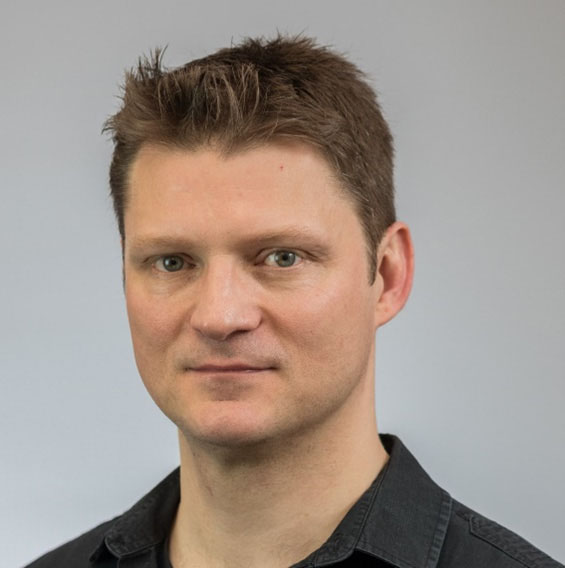}}]
  {Nathan F. Lepora} received the B.A. degree in Mathematics and the Ph.D. degree in Theoretical Physics from the University of Cambridge, Cambridge, U.K. He is currently a Professor of Robotics and AI with the University of Bristol, Bristol, U.K. He leads the Dexterous Robotics Group in Bristol Robotics Laboratory. Prof. Lepora is a recipient of a Leverhulme Research Leadership Award on ‘A Biomimetic Forebrain for Robot Touch’. He coedited the book “Living Machines” that won the 2019 BMA Medical Book Awards (basic and clinical sciences category). His research team won the ‘University Research Project of the Year’ at the 2022 Elektra Awards.

\end{IEEEbiography}

\end{document}